\documentclass{article}
\usepackage{PRIMEarxiv}

\usepackage[utf8]{inputenc} 
\usepackage[T1]{fontenc}    
\usepackage{url}            
\usepackage{booktabs}       
\usepackage{amsfonts}       
\usepackage{nicefrac}       
\usepackage{microtype}      
\usepackage{lipsum}
\usepackage{graphicx}

\usepackage{algorithm}
\usepackage{algpseudocode}

\usepackage{colortbl}
\usepackage[dvipsnames,table,xcdraw]{xcolor}

\definecolor{lightgray}{gray}{0.8} 

\usepackage{amsmath}

\usepackage{amssymb}
\usepackage{braket}
\usepackage{placeins}
\usepackage{makecell}
\usepackage{multirow}
\usepackage{caption}
\usepackage{moresize}

\newif\ifARXIV
\newif\ifIOP
\ARXIVtrue
\IOPfalse 
\newif\ifElsavier
\ARXIVfalse
\Elsaviertrue 

\title{
Learning Fluid-Structure Interaction with Physics-Informed Machine Learning and Immersed Boundary Methods
}

\author{
  Afrah Farea$^{1}$, Saiful Khan$^{2}$, Reza Daryani$^{1}$, Emre Cenk Ersan$^{1}$, Mustafa Serdar Celebi$^{1}$ \\[0.5em]
  $^{1}$Computational Science and Engineering Department, Informatics Institute \\
  Istanbul Technical University, Istanbul 34469, Turkiye \\
  \texttt{farea16@itu.edu.tr, daryani@itu.edu.tr, ersane@itu.edu.tr, mscelebi@itu.edu.tr} \\[0.5em]
  $^{2}$Scientific Computing, Rutherford Appleton Laboratory \\
  Science and Technology Facilities Council (STFC), OX11 0QX, United Kingdom\\
  \texttt{saiful.khan@stfc.ac.uk}  \\
}

\begin{document}
\maketitle

\begin{abstract}
Physics-informed neural networks (PINNs) have emerged as a promising approach for solving complex fluid dynamics problems, yet their application to fluid-structure interaction (FSI) problems with moving boundaries remains largely unexplored. This work addresses the critical challenge of modeling FSI systems with moving interfaces, where traditional unified PINN architectures struggle to capture the distinct physics governing fluid and structural domains simultaneously.
We present an innovative Eulerian-Lagrangian PINN architecture that integrates immersed boundary method (IBM) principles to solve FSI problems with moving boundary conditions. Our approach fundamentally departs from conventional unified architectures by introducing domain-specific neural networks: an Eulerian network for fluid dynamics and a Lagrangian network for structural interfaces, coupled through physics-based constraints. Additionally, we incorporate learnable B-spline activation functions with SiLU to capture both localized high-gradient features near interfaces and global flow patterns.
Empirical studies on a 2D cavity flow problem involving a moving solid structure show that while baseline unified PINNs achieve reasonable velocity predictions, they suffer from substantial pressure errors (12.9\%) in structural regions. Our Eulerian-Lagrangian architecture with learnable activations (EL-L) achieves better performance across all metrics, improving accuracy by 24.1-91.4\% and particularly reducing pressure errors from 12.9\% to 2.39\%.
These results demonstrate that domain decomposition aligned with physical principles, combined with locality-aware activation functions, is essential for accurate FSI modeling within the PINN framework. 
\end{abstract}

\section{Introduction}

Fluid Structure Interaction (FSI) phenomena govern critical processes across engineering disciplines, from blood flow through heart valves in biomedicine to wing flutter in aerospace systems  \cite{dowell2001modeling, hughes2013aerospace, liu2017stability, mao2019water, nieva2012fluid, esposito2014monitoring, hirschhorn2020fluid, syed2023modeling, wang2008combined, wang2012fluid,xie2016perfectly, dai2021numerical}. 
These problems involve complex bidirectional coupling: fluid forces deform structures, which in turn alter flow patterns, creating a nonlinear feedback loop that challenges conventional computational methods. While computational fluid dynamics (CFD) coupled with structural solvers can achieve high-fidelity results, the computational cost of resolving fine-scale boundary movements and interfacial forces often renders these approaches impractical for real-time applications or parametric studies.

The immersed boundary method (IBM) has emerged as an efficient alternative for solving FSI simulations by embedding moving structures within fixed Eulerian grids, eliminating costly remeshing operations. 
Despite these advantages, the method suffers from fundamental limitations that restrict its broader applicability.
Enforcing interfacial coupling conditions, e.g., no-slip and pressure continuity, can be challenging, particularly at high Reynolds numbers or significant structural deformations~\cite{tschisgale2020immersed}.
The method's reliance on regularized delta functions introduces numerical diffusion near interfaces, compromising accuracy precisely where it matters most~\cite{griffith2009simulating}. Furthermore, IBM implementations typically require problem-specific parameter tuning, limiting their generalizability across different geometries and flow regimes~\cite{bhalla2013unified, griffith2017hybrid}.

Physics Informed Neural Networks (PINNs) offer a compelling approach by encoding governing equations directly into neural network loss functions~\cite{han2022deep, zhang2023airfoil, liu2024novel, xiao2024fourier, fan2024differentiable, mazhar2024new, gao2024predicting, farea2025:learnable,jia2024flow, farea2025:qcpinn}, potentially addressing IBM's limitations through their mesh-free nature and automatic differentiation capabilities.

Recent work has demonstrated PINNs' effectiveness for stationary FSI problems, with IB-PINN frameworks~\cite{fang2022immersed, huang2022direct, sundar2024physics} successfully enforcing interface conditions through penalty methods. However, these approaches remain constrained to steady flows and fixed geometries. Studies incorporating operator learning~\cite{xiao2024fourier} and reduced-order modeling~\cite{han2022deep} have tackled more complex dynamics but sacrifice explicit physics enforcement and require extensive training data. Critically, no existing work addresses the fundamental challenge of FSI problems with moving and deformable boundaries.

This heterogeneity presents a core architectural challenge: unified neural networks struggle to simultaneously optimize for smooth velocity fields in the fluid domain and sharp pressure gradients at moving interfaces. Our statistical analysis of FSI systems reveals that pressure variability in the case study near the structure interface significantly exceeds that in bulk fluid regions, while velocity fields exhibit opposite trends. This fundamental mismatch suggests that effective PINN architectures for FSI must align with the underlying physics rather than treating all domains uniformly.
We address this challenge through two key contributions. 

\begin{enumerate}
    \item
    Inspired by the IBM method, we introduce a decoupled Eulerian-Lagrangian architecture that maintains separate neural networks for fluid and structural domains, enabling each to specialize in its respective physics while coupling through interface constraints.

    \item 
    We incorporate learnable B-spline basis functions enhanced with SiLU to provide locality-aware representation, which is crucial for capturing high-gradient features near moving boundaries while maintaining smooth approximation in bulk regions.

\end{enumerate}

We systematically evaluate four architectural variants: baseline and Eulerian-Lagrangian configurations, each with fixed Tanh and learnable B-spline+SiLU activation functions. 
We trained and evaluated the models using a dataset generated from a simulation with a moving object modeled by IBM, as explained in~\cite{griffith2017hybrid}. We verified the simulation against high-precision numerical solutions obtained from the Immersed Boundary Adaptive Mesh Refinement (IBAMR) simulation software~\cite{ibamr}. 
The dataset, source code, and pre-trained models from this work are publicly available at \url{https://github.com/afrah/pinn_fsi_ibm}~\cite{afrah2024pinn_ibm4fsi}.


\section{Related Work}

Recent years have seen growing interest in leveraging neural networks to solve FSI problems, particularly through integrating physics-informed machine learning approaches with immersed boundary techniques.

\noindent\textbf{Early PINN-IBM developments.} Fang et al.~\cite{fang2022immersed} developed an immersed boundary-PINN (IB-PINN) framework for fluid-solid coupling, demonstrating promising results in modeling 2D flow past a static cylinder. However, their model was restricted to stationary solid boundaries throughout the domain. Building on this foundation, Huang et al.~\cite{huang2022direct} proposed a direct-forcing immersed boundary PINN that introduced velocity and force penalty terms to enforce interface conditions. While effective for steady flows around fixed geometries, their formulation does not support deforming or moving structures.

\textbf{Operator learning approaches.} Xiao et al.~\cite{xiao2024fourier} employed a Fourier Neural Operator-based approach to predict vesicle dynamics in fluids, offering efficiency in long-term dynamics and parameter generalization. However, their operator learning approach lacks physics-constrained enforcement of interfacial conditions and is less suited for tasks requiring high-fidelity resolution of boundary dynamics.

\textbf{Reduced-order modeling.} Han et al.~\cite{han2022deep} introduced a neural network-based reduced-order model (ROM) for vortex-induced vibration (VIV) problems. While reduced-order models offer efficient surrogate approximations, they require extensive precomputed high-fidelity data and lack the physics-consistency and flexibility of PINNs in extrapolating to new scenarios or partial observations.

\textbf{Moving boundary frameworks.} Sundar et al.~\cite{sundar2024physics} explored PINN-based frameworks for moving boundary problems by proposing two variants: MB-PINN, which considers only the fluid region, and MB-IBM-PINN, which includes both fluid and solid regions. They investigated relaxation of physics constraints and region-specific loss weighting to improve accuracy and generalization. Their work highlights key challenges in training PINNs for moving boundary problems, particularly regarding pressure recovery and accurate resolution of localized flow features.


\section{FSI with IBM}

\begin{table*}[tb]
\caption{Statistical summary for FSI-IBM model problem with velocity and pressure values. While velocity fields show relatively lower variability, pressure exhibits the highest variability at the structure interface.
}
\centering

\small 

\begin{tabular}
    {>{\raggedright\arraybackslash}m{2cm}
    c>{\centering\arraybackslash}m{2cm}
    >{\centering\arraybackslash}m{7cm}}
    
    \hline
    Domain & Field & Standard Deviation & Distribution \\
    \hline
            
    \multirow{7}{*}{Fluid}& $u$&  0.208 & \includegraphics[width=6.0cm,trim=0.0cm 0.2cm 0.2cm 0.3cm, clip]{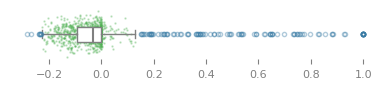}\\[2ex]
   
    & $v$&  0.130 & \includegraphics[width=6.0cm,trim=0.0cm 0.2cm 0.2cm 0.3cm, clip]{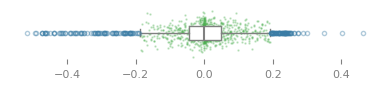}\\[2ex]

    & $p$& 0.115 & \includegraphics[width=6.0cm,trim=0.0cm 0.2cm 0.2cm 0.3cm, clip]{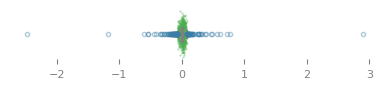}\\[2ex]

    \noalign{\vskip 1ex}

    
    \multirow{6}{*}{\makecell{Structure }}&$u$&  0.145 &  \includegraphics[width=6.0cm,trim=0.0cm 0.2cm 0.2cm 0.3cm, clip]{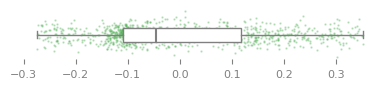}\\[2ex]
    
    & $v$& 0.133 &\includegraphics[width=6.0cm,trim=0.0cm 0.2cm 0.2cm 0.3cm, clip]{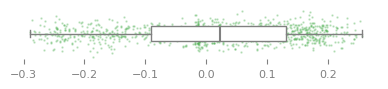}\\[2ex]

    & $p$&  0.181 &\includegraphics[width=6.0cm,trim=0.0cm 0.2cm 0.2cm 0.3cm, clip]{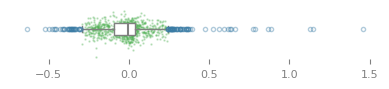}\\[2ex]
    
    \hline

\end{tabular}
\label{tab:ibm_dist}

\end{table*}

\begin{figure}[t]
    \centering
    \includegraphics[width=1.0\textwidth,trim=0.0cm 0.0cm 0.0cm 0.0cm, clip]{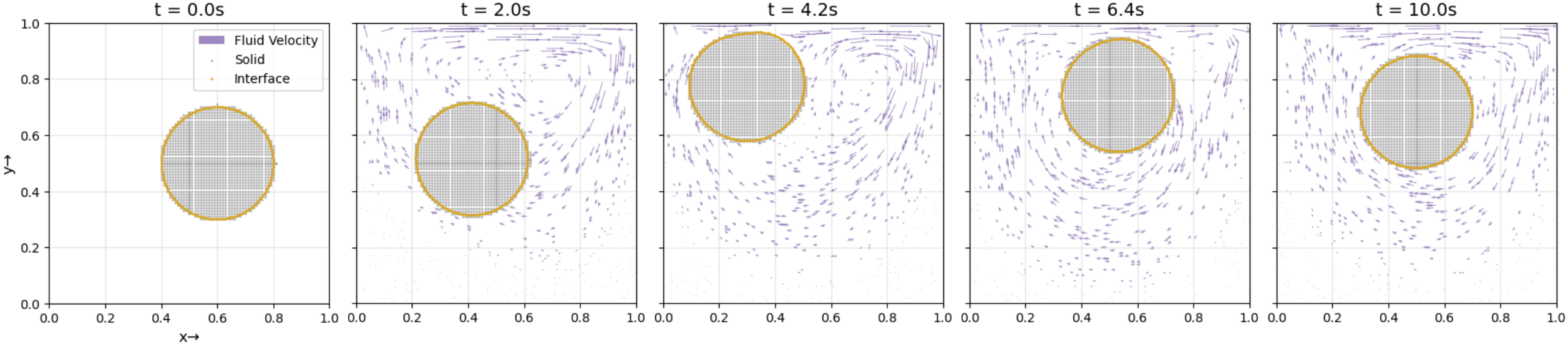}    
    \caption{
    Illustration of the computational domain of the FSI problem considered in the present work, showing the movement of a soft disc in a lid-driven cavity flow at different time steps.
     }
    \label{fig:problem_domain}
\end{figure}

FSI involves complex bidirectional coupling where fluid forces deform structures, which in turn alter flow patterns, creating nonlinear feedback loops that challenge conventional computational methods. 
To address these challenges, various numerical strategies have been developed to solve the FSI problem. These methods can be classified into body-conforming or non-body-conforming based on whether the fluid mesh aligns with the structure. 
Body-fitted, mesh-conforming approaches include Arbitrary Lagrangian-Eulerian (ALE) formulations and finite-element or finite-volume methods. Non-body-conforming methods encompass immersed boundary methods and fictitious domain approaches, with specialized applications using mesh-free particle techniques and reduced-order models.

 Among these approaches, IBM emerges as an efficient alternative by embedding moving structures within fixed Eulerian grids, eliminating costly remeshing operations. 
 IBM employs a static Cartesian mesh for the fluid region and curvilinear Lagrangian nodes for structural domains that move freely over the background mesh. 
 The grids for the movable structures do not need to conform to the fixed grid of the fluid, allowing the IBM to impose coupling conditions more effectively on the fluid-structure interface.
 This characteristic makes IBM particularly suitable for significant displacement problems, such as modeling biological valves~\cite{griffith2009simulating,griffith2012immersed}.

The coupling between Eulerian and Lagrangian variables can be achieved using discrete~\cite{uhlmann2005immersed} or continuous~\cite{peskin2002immersed} forcing schemes. Discrete forcing applies boundary conditions using grid cells in the solid region with interpolation schemes to enforce interface conditions. Continuous forcing utilizes compact delta functions at the interface to facilitate quantity transfer between fluid and immersed structures. This work adopts continuous forcing due to its advantage of representing fluid-structure interfaces through evenly distributed surface marker points without requiring special grid point identification.

\subsection{Problem Setup and Assumptions}

This study examines an FSI scenario involving a moving circular elastic solid object within a fluid, using a direct forcing IBM. Following the setup described in~\cite{griffith2017hybrid}, we model the fluid as a time-dependent, two-dimensional cavity flow discretized with finite differences. Lagrangian marker points represent the solid boundary, while the solid domain is discretized using finite elements and no-slip boundary conditions at the solid boundary.

The domain $\Omega$ $ = [0, 1] \times [0, 1]$ is a two-dimensional square cavity with uniform discretization $(N_x, N_y) = (100, 100)$. The immersed structure is a disc with a radius of 0.2, initially centered at $\mathbf{x}= ( 0.6, 0.5)$. The time span $0 \leq t \leq 10$, with $\Delta t = 0.01$, during which the disc completes slightly more than one full rotation.
The boundary $\Gamma_1$ represents the top Dirichlet boundary with tangential velocity, while $\Gamma_0$ denotes the three stationary sides. The Reynolds number is $Re=100$. During the simulation, the flow induced by the driven lid causes the structure to come nearly into contact with the moving upper boundary of the domain, as illustrated in Fig.~\ref{fig:problem_domain}.

Table~\ref{tab:ibm_dist} summarizes statistical variability of velocity and pressure fields across domains. 
Fluid variables exhibit relatively low variability with smooth distributions, while the fluid-structure interface shows substantially higher variability, particularly in pressure ($\sigma_p = 0.181$), highlighting sharp gradients and localized dynamics at the moving interface.
These observations directly motivated the development of our proposed architecture.

The governing equations for the fluid domain consist of the incompressible Navier-Stokes equations with coupling conditions:

\ifElsavier
\begin{subequations}\label{eq:IBM}
\footnotesize
    \text{\textbf{Navier-Stokes Equations:}}
  \begin{align}
    \rho_f \left(\frac{\partial \mathbf{u}_f(t,\mathbf{x})}{\partial t} + \mathbf{u}_f(t,\mathbf{x}) \cdot \nabla\mathbf{u}_f(t,\mathbf{x})\right) &= 
        -\nabla p_f(\mathbf{x},t) + \mu_f \nabla^2 \mathbf{u}_f(t,\mathbf{x}) + \mathbf{f}_{\text{e}}(t,\mathbf{x}), 
        \quad \text{in } \Omega_f \label{eq:IBM_d} \\
    \nabla \cdot \mathbf{u}_f(t,\mathbf{x}) &= 0, \quad \text{in } \Omega_f  \notag\\
    \mathbf{u}(0,\mathbf{x}) &= 0, \quad \text{in } \Omega \notag\\
    \mathbf{u}(t,\mathbf{x}_0) &= 0, \quad \text{in } \Gamma_i \notag\\
    \mathbf{u}(t,\mathbf{x}_1) &= 1, \quad \text{in } \Gamma_0 \notag
  \end{align}
  \text{\textbf{Fluid Structure Coupling Equations (Interface Condition):}}
  \begin{align}
    \frac{\partial \mathbf{d}_s(\xi(t, s))}{\partial t} &= \mathbf{u}_s(\xi(t, s)) = 
        \int_\Omega \mathbf{u}_f(t,\mathbf{x})\delta(\mathbf{x} - \xi(t, s)) \, d\mathbf{x}  = \mathbf{u}_f(\xi(t, s)), & \text{in } \Omega_f \cap \Omega_s   \label{eq:IBM_couple_velocity} \\
          \nabla p(\xi(t, s)) \cdot \mathbf{n} &= n_x* \frac{\partial p(\xi(t, s))}{\partial x } + n_y* \frac{\partial p(\xi(t, s))}{\partial y }=0 & \text{in } \Omega_f \cap \Omega_s \label{eq:IBM_pressure_gradient}\\
    \mathbf{f}_{\text{e}}(t , \mathbf{x}) &= \int_{\Gamma_s} \mathbf{F}_{\text{l}} (\xi(t , s)) \delta(\mathbf{x} - \xi(t ,s)) \, ds \label{eq:IBM_couple_force} & \text{in } \Omega_f \cap \Omega_s \\
    &\approx \sum_{i} \mathbf{F}_{\text{l}}(\xi(t , s_i)) \delta_h(\mathbf{x} - \xi(t ,s_i)) \Delta s_i \notag \\
    &\approx \sum_{i} \mathbf{F}_{\text{l}}(\xi(t ,s_i)) \frac{1}{2\pi \sigma^2} \exp\left(-\frac{\|\mathbf{x} - \xi(t ,s_i)\|^2}{2\sigma^2}\right) \Delta s_i \notag 
  \end{align}
\end{subequations}

\fi
\ifARXIV
\begin{subequations}\label{eq:IBM}
    \text{\textbf{Navier-Stokes Equations:}}
  \begin{align}
    \rho_f \left(\frac{\partial \mathbf{u}_f(t,\mathbf{x})}{\partial t} + \mathbf{u}_f(t,\mathbf{x}) \cdot \nabla\mathbf{u}_f(t,\mathbf{x})\right) &= 
        -\nabla p_f(\mathbf{x},t) + \mu_f \nabla^2 \mathbf{u}_f(t,\mathbf{x}) + \mathbf{f}_{\text{e}}(t,\mathbf{x}), 
        \quad \text{in } \Omega_f \label{eq:IBM_d} \\
    \nabla \cdot \mathbf{u}_f(t,\mathbf{x}) &= 0, \quad \text{in } \Omega_f  \notag\\
    \mathbf{u}(0,\mathbf{x}) &= 0, \quad \text{in } \Omega \notag\\
    \mathbf{u}(t,\mathbf{x}_0) &= 0, \quad \text{in } \Gamma_i \notag\\
    \mathbf{u}(t,\mathbf{x}_1) &= 1, \quad \text{in } \Gamma_0 \notag
  \end{align}
  \text{\textbf{Fluid Structure Coupling Equations (Interface Condition):}}
  \begin{align}
    \frac{\partial \mathbf{d}_s(\xi(t, s))}{\partial t} &= \mathbf{u}_s(\xi(t, s)) = 
        \int_\Omega \mathbf{u}_f(t,\mathbf{x})\delta(\mathbf{x} - \xi(t, s)) \, d\mathbf{x}  = \mathbf{u}_f(\xi(t, s)), & \text{in } \Omega_f \cap \Omega_s   \label{eq:IBM_couple_velocity} \\
          \nabla p(\xi(t, s)) \cdot \mathbf{n} &= n_x* \frac{\partial p(\xi(t, s))}{\partial x } + n_y* \frac{\partial p(\xi(t, s))}{\partial y }=0 & \text{in } \Omega_f \cap \Omega_s \label{eq:IBM_pressure_gradient}\\
    \mathbf{f}_{\text{e}}(t , \mathbf{x}) &= \int_{\Gamma_s} \mathbf{F}_{\text{l}} (\xi(t , s)) \delta(\mathbf{x} - \xi(t ,s)) \, ds \label{eq:IBM_couple_force} & \text{in } \Omega_f \cap \Omega_s \\
    &\approx \sum_{i} \mathbf{F}_{\text{l}}(\xi(t , s_i)) \delta_h(\mathbf{x} - \xi(t ,s_i)) \Delta s_i \notag \\
    &\approx \sum_{i} \mathbf{F}_{\text{l}}(\xi(t ,s_i)) \frac{1}{2\pi \sigma^2} \exp\left(-\frac{\|\mathbf{x} - \xi(t ,s_i)\|^2}{2\sigma^2}\right) \Delta s_i \notag 
  \end{align}
\end{subequations}
\fi
\noindent where $p_f$ is a scalar pressure field, $\mathbf{f}_{\text{e}}(t,\mathbf{x})$ is the Eulerian elastic force density exerted by the solid on the fluid and distributed using the immersed boundary method.  $\mathbf{F}_{\text{l}}$ is the Lagrangian elastic force density added to the solid interface. $\mathbf{u(t , x)} = (u(t , x), v(t , y))$ is the velocity field, where $\mathbf{x} = (x, y)$, $\xi(t ,s)$ is the embedded solid interface representing the Lagrangian points,  $\mu_f$ is the dynamic viscosity of the fluid such that $\mu_f=\frac{1.0}{\text{Re}} = 0.01$, with Re as the Reynold number, $\rho_f$  represents the fluid density, set to unity.

To compute the normal vector $\mathbf{n}$ at the interface, we use the outward normal by rotating the tangent vector by $90^\circ$ counterclockwise and normalizing:
\begin{align*}
\mathbf{n}_i =
\frac{(-t_{y,i}, \; t_{x,i})}{\sqrt{t_{x,i}^2 + t_{y,i}^2 + \epsilon}},
\end{align*}

\noindent where $t_i$ is the tangent at each point, approximated by central differences with periodic indexing:

\begin{align*}
\mathbf{t}_i = \bigl(x_{(i+1)\bmod N} - x_{(i-1)\bmod N}, \;
                   y_{(i+1)\bmod N} - y_{(i-1)\bmod N}\bigr).
\end{align*}
\noindent where $\{\mathbf{x}_i\}_{i=0}^{N-1}$ denote the ordered marker points along the closed interface.

The coupling conditions enforce no-slip boundary conditions (Eq.~\ref{eq:IBM_couple_velocity}) and pressure gradient continuity (Eq.~\ref{eq:IBM_pressure_gradient}) at the fluid-structure interface. We adopt the ``zero-thickness assumption''~\cite{peskin2002immersed}, where the structural domain is represented solely by its boundary without resolving finite thickness. We also relax the standard IBM force-spreading equation, relying instead on Eq.~\ref{eq:IBM_couple_velocity} and~\ref{eq:IBM_pressure_gradient} to enforce fluid-structure coupling.


\begin{figure}[t]
    \centering
    \includegraphics[width=0.6\textwidth,trim=0.0cm 0.0cm 0cm 0.0cm, clip]{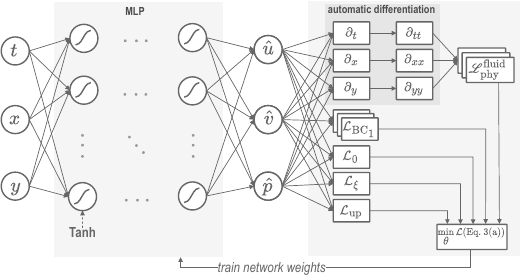}     
    \caption{
    Baseline (B): Standard PINN with fixed Tanh activation. 
    $\{t, x, y\}$ as inputs and output $\{\hat{u}, \hat{v}, \hat{p}\}$ for velocity components and pressure in both Eulerian and Lagrangian domains. 
    The loss function $\mathcal{L}(\theta)$ is defined in Eq.~\ref{equ:m1_main_loss}. 
    }
    \label{fig:B}
\end{figure}

\begin{figure}[ht]
    \centering
    \includegraphics[width=0.6\textwidth,trim=0.0cm 0.0cm 0cm 0.0cm, clip]{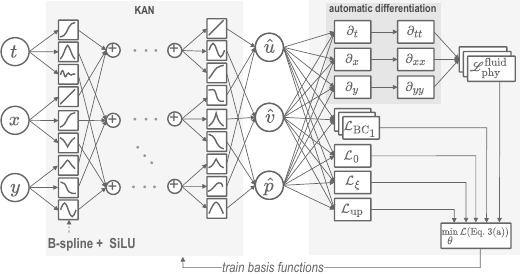}     
    \caption{
    Baseline enhanced with learnable activation (B-L) PINN, which includes trainable B-spline+SiLU activation functions. 
    Both models take $\{t, x, y\}$ as inputs and output $\{\hat{u}, \hat{v}, \hat{p}\}$ for velocity components and pressure in both Eulerian and Lagrangian domains. 
    The loss function $\mathcal{L}(\theta)$ is defined in Eq.~\ref{equ:m1_main_loss}. 
    }
    \label{fig:B-L}
\end{figure}

\begin{figure}[ht]
    \centering
    \includegraphics[width=0.6\textwidth,trim=0.0cm 0.0cm 0cm 0.0cm, clip]{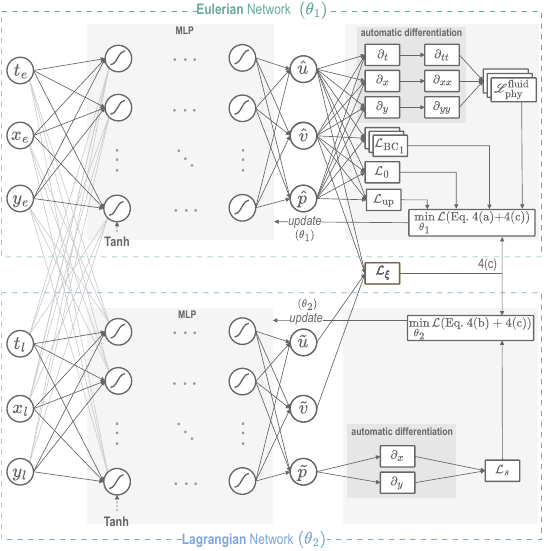}    
    \caption{Eulerian-Lagrangian network: the inputs of the Eulerian Network ($\theta_1$) are $\{t_e, x_e, y_e\}$ from the Eulerian fluid domain, and the output variables are $\{\hat{u}, \hat{v},\hat{p}\}$ representing the Eulerian velocity and pressure fields respectively. For the Lagrangian Network ($\theta_2$), the input variables are $\{t_l, x_l, y_l\}$ representing the Lagrangian structure and the output variables are $\{\Tilde{u}, \Tilde{v}, \Tilde{p}\}$ representing the Lagrangian velocity and pressure values respectively. The inputs are also shared between the networks for the velocity prediction at the interface.
     See, Eq.\ref{eq:m2_main_loss} for details of the loss function $\mathcal{L}(\theta)$.
    }
    \label{fig:EL}
\end{figure}

\begin{figure}[htb]
    \centering
    \includegraphics[width=0.6\textwidth,trim=0.0cm 0.0cm 0cm 0.0cm, clip]{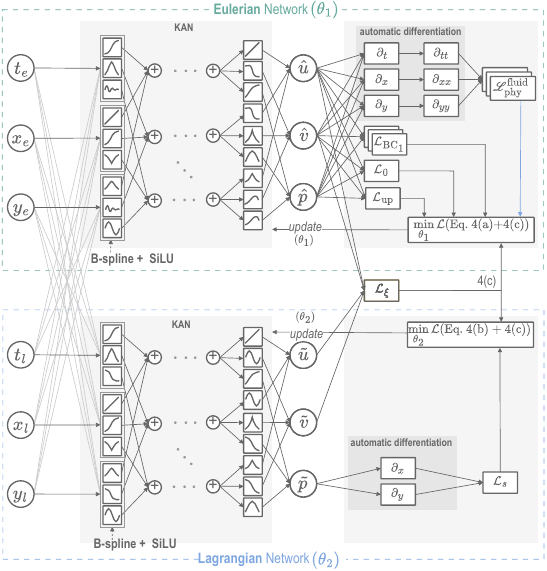}    
    \caption{Eulerian-Lagrangian network: the inputs of the Eulerian Network ($\theta_1$) are $\{t_e, x_e, y_e\}$ from the Eulerian fluid domain, and the output variables are $\{\hat{u}, \hat{v},\hat{p}\}$ representing the Eulerian velocity and pressure fields respectively. For the Lagrangian Network ($\theta_2$), the input variables are $\{t_l, x_l, y_l\}$ representing the Lagrangian structure and the output variables are $\{\Tilde{u}, \Tilde{v}, \Tilde{p}\}$ representing the Lagrangian velocity and pressure values respectively. The inputs are also shared between the networks for the velocity prediction at the interface.
    See Eq.\ref{eq:m2_main_loss} for details of the loss function $\mathcal{L}(\theta)$.
    }
    \label{fig:EL-L}
\end{figure}

\section{Neural Network Architectures}
\label{sec:proposed_networks}

We implement four neural network architectures, including one baseline PINN architecture and three improved architectures, to systematically evaluate our proposed solution.
This section details their networks and their loss function design.

\subsection{Baseline (B)}
\label{sec:b}

The baseline architecture employs a standard PINN with a fixed Tanh activation function, implemented as a fully connected multi-layer perceptron (MLP) as illustrated in Fig.~\ref{fig:B}. This unified design takes $\{t,x,y\}$ as inputs from both Eulerian and Lagrangian domains and predicts the velocity components and pressure values, $\{\hat{u},\hat{v},\hat{p}\}$ across the entire computational domain within a shared parameter space.

The complexity of FSI systems requires careful balancing of multiple physics constraints through weighted loss function terms. The overall loss function, $L(\theta)$, combines governing equations and boundary conditions:

\begin{subequations}
\footnotesize
    \begin{align}
\mathcal{L}(\theta) &=  \lambda_1 \|\mathcal{L}_{\text{phy}}^{\text{fluid}}\|_{ \Omega_f}  +  \lambda_2 \big( \|\mathcal{L}_{\text{up}}\|_{ \Gamma_1}   + \|\mathcal{L}_{\text{bc}_1}\|_{ \Gamma_0} \big) +  \lambda_3 \|\mathcal{L}_0\|_{\Omega_0}   + \lambda_4 \|\mathcal{L}_{\xi}\|_{\xi}, 
    \label{equ:m1_main_loss}
\intertext{where $\mathcal{L}_{\text{phy}}^{\text{fluid}} =\mathcal{L}^{\text{fluid}}_{r_u} +\mathcal{L}^{\text{fluid}}_{r_v}+ \mathcal{L}^{\text{fluid}}_{r_c}$. We select weights $\lambda_1=0.1, \lambda_2=2.0, \lambda_3=4.0, \lambda_4=0.1$ based on empirical optimization.}
    \mathcal{L}^{\text{fluid}}_{r_u}(\theta) &= \text{MSE}\left[ \frac{\partial \hat{u}}{\partial t} + \hat{u} \frac{\partial \hat{u}}{\partial x} + \hat{v} \frac{\partial \hat{u}}{\partial y} + \frac{1.0}{ \rho_{f}} \frac{\partial \hat{p}}{\partial x} - \mu  \left(\frac{\partial^2 \hat{u}}{\partial x^2} + \frac{\partial^2 \hat{u}}{\partial y^2}\right) \right],  \label{eq:m1_phy_ru}\\
    \mathcal{L}^{\text{fluid}}_{r_v}(\theta) &=\text{MSE}\left[\frac{\partial \hat{v}}{\partial t} + \hat{u} \frac{\partial \hat{v}}{\partial x} + \hat{v} \frac{\partial \hat{v}}{\partial y} + \frac{1.0}{ \rho_{f}}\frac{\partial \hat{p}}{\partial y} - \mu \left(\frac{\partial^2 \hat{v}}{\partial x^2}+ \frac{\partial^2 \hat{v}}{\partial y^2}\right) \right], \label{eq:m1_phy_rv}\\
    \mathcal{L}^{\text{fluid}}_{r_c}(\theta) &= \text{MSE}\left[\frac{\partial \hat{u}}{\partial x} + \frac{\partial \hat{v}}{\partial y}\right], \label{eq:m1_phy_rc}\\
    \mathcal{L}_{\text{up}}(\theta) &= \text{MSE}\left[1.0 - \hat{u}\right] + \text{MSE}\left[ \hat{v}\right], \label{eq:m1_loss_up}\\ 
    \mathcal{L}_{\text{bc}_1}(\theta) &= \mathcal{L}_{\text{bottom, right, left}} = \text{MSE}\left[ \mathcal{L}(\hat{u}) + \mathcal{L}( \hat{v})\right], \label{eq:m1_loss_bc1}\\ 
    \mathcal{L}_0(\theta) &= \text{MSE}\left[ \hat{u} + \hat{v} + \hat{p}\right], \label{eq:m1_loss_init}\\
    \mathcal{L}_{\xi}(\theta) &=  \text{MSE}\left[ n_x\frac{\partial \hat{p}(\xi(s,t))}{\partial x} + n_y\frac{\partial \hat{p}(\xi(s,t))}{\partial y} \right] \label{eq:velocity_xi}
    \end{align}
\end{subequations}
\noindent where $\mathcal{L}_{phy}^{fluid}$, $\mathcal{L}_{s}$  are the losses of the fluid and structure, respectively. $\mathcal{L}_{left} $,  $\mathcal{L}_{right}$, $\mathcal{L}_{up}$,  $\mathcal{L}_{bottom}$,  $\mathcal{L}_{{u}_0}$, are the left, right, up, bottom and initial losses respectively.

\subsection{Baseline with Learnable Activation (B-L)}
\label{sec:B-L}

This architecture enhances the baseline by replacing fixed Tanh activations with learnable B-spline combined with SiLU functions, implemented using Kolmogorov-Arnold Networks (KAN)~\cite{liu2024kan} as shown in Fig.~\ref{fig:B-L}. This modification enables dynamic adaptation to localized flow variations near boundaries and fluid-structure interfaces while maintaining the unified parameter space.
The hybrid activation function combines local adaptability with global expressivity:

\begin{align}
\label{bspline_activation}
\phi(x) &= \lambda_0 \cdot \text{SiLU}(x) + \sum_{i=1}^{d+k-1} c_i B_i^d(x),
\end{align}
where $d=3$ (cubic splines), $k=8$ (grid intervals), and $c_i, \lambda_0 \in \mathbb{R} $ are learnable parameters.

The B-splines $B_i^d(x) $ are defined recursively using the Cox-de Boor formula~\cite{de1972calculating}:

\begin{align*}
B_i^d(x) &= \frac{x - \xi_i}{\xi_{i+d} - \xi_i} B_i^{d-1}(x) + \frac{\xi_{i+d+1} - x}{\xi_{i+d+1} - \xi_{i+1}} B_{i+1}^{d-1}(x)
\end{align*}
\noindent with the base case ($d = 0$):

\begin{align*}
B_i^0(x) &= \begin{cases}1, & \text{if } \xi_i \leq x < \xi_{i+1} \\0, & \text{otherwise}\end{cases}
\end{align*}
\noindent where $\{\xi_i\} $ are the knot points that define the spline, which is dynamically updated during training based on the distribution of input values.

The B-splines provide local adaptability through compact support and learnable control points, while SiLU enhances global expressivity with smooth nonlinearity. This combination enables precise adjustments in high-complexity regions while maintaining stable training and capturing overall flow patterns.

For three inputs ($n = 3$), $\mathbf{x} = (t, x, y) \in \mathbb{R}^3 $, the output of our network is computed as:

\begin{align}
\label{kan_output}
f(\mathbf{x}) &= f(t, x, y) = \sum_{i=1}^{2n+1} \psi_i \left( \sum_{j=1}^{n} \phi_{ij}(x_j) \right),
\end{align}
where $\psi_i: \mathbb{R} \rightarrow \mathbb{R} $ are the outer transformations, and $\phi_{ij}: \mathbb{R} \rightarrow \mathbb{R} $ are the KAN activations applied to each input component.
Substituting for Eq.~\ref{bspline_activation}, Eq.~\ref{kan_output} yields:
\begin{align*}
f(\mathbf{x}) &= f_L \circ f_{L-1} \circ \cdots \circ f_1 (\mathbf{x})
\end{align*}
where each layer $f_\ell$ applies the transformation:

\begin{align*}
f_\ell(\mathbf{z}) &= \sum_{i=1}^{n_\ell} \left( \sum_{j=1}^{n_{\ell-1}} \left( \lambda^{(0)}_{ij} \cdot \text{SiLU}(z_j) + \sum_{r=1}^{g+d-1} c_{ijr} \cdot B_r^d(z_j) \right) \right)
\end{align*}
\noindent with $n_\ell$ being the number of neurons in layer $\ell$, $g$ being the grid size, and $d$ being the spline order.

\subsection{Eulerian-Lagrangian (EL)}
\label{sec:EL}

Our statistical analysis in Table~\ref{tab:ibm_dist} reveals substantial variability in target variables across domains, with particularly large disparities between the fluid and structure regions. 
This heterogeneity challenges unified architectures that force incompatible physics into shared parameter spaces. 
Inspired by IBM's natural separation of Eulerian fluid and Lagrangian structural representations, we propose a decoupled architecture with two specialized subnetworks (as in Fig.~\ref{fig:B})).

\begin{itemize}
    \item 
    \textbf{Eulerian network ($\theta_1$)}: Handles fluid dynamics with inputs $\{t_e, x_e, y_e\}$ and outputs $\{\hat{u}, \hat{v}, \hat{p}\}$ 
    \item 
    \textbf{Lagrangian network ($\theta_2$)}: Manages structural interfaces with inputs $\{t_l, x_l, y_l\}$ and outputs $\{\tilde{u}, \tilde{v}, \tilde{p}\}$
\end{itemize}
    
This specialization minimizes the spectral bias of the neural network models and enables each network to optimize for domain-specific statistical distributions and physics without parameter interference, while coupling occurs through interface constraints that enforce no-slip conditions and pressure continuity.

The decoupled loss function separates domain-specific and coupling terms:

\begin{subequations}
\footnotesize
    \begin{align}
        \mathcal{L}(\theta_1) &=   \lambda_1 \|\mathcal{L}_{\text{phy}}^{\text{fluid}}\|_{ \Omega_f} + \lambda_2 \left( \|\mathcal{L}_{\text{up}}\|_{ \Gamma_1}  + \|\mathcal{L}_{\text{bc}_1}\|_{ \Gamma_0} \right) +  \lambda_3 \|\mathcal{L}_{u_0}\|_{\Omega}  \label{eq:m2_main_loss}  \\
         \mathcal{L}(\theta_2) &= \lambda_4 \mathcal{L}_{s} \\
         \mathcal{L}(\theta_1 \cup \theta_2) &=   \lambda_5 \mathcal{L}_{\xi}  \label{eq:m3_coupling}\\
        \intertext{where $\mathcal{L}_{\text{phy}}^{\text{fluid}} = \mathcal{L}^{\text{fluid}}_{r_u} +\mathcal{L}^{\text{fluid}}_{r_v}+ \mathcal{L}^{\text{fluid}}_{r_c}$. Where weights are $\lambda_1=0.1, \lambda_2=4.0, \lambda_3=1.0, \lambda_4=0.1, \lambda_5=0.1$.}
        \mathcal{L}_{\text{up}}(\theta_1) &= \text{MSE}\left[ 1.0 - \hat{u} \right] + \text{MSE}\left[ \hat{v} \right] \\ 
        \mathcal{L}_{\text{bc}_1}(\theta_1) &= \mathcal{L}_{\text{bottom, right, left}} = \text{MSE}\left[ \mathcal{L}(\hat{u}) + \mathcal{L}( \hat{v}) \right] \\ 
        \mathcal{L}_{u_0}(\theta_1) &= \text{MSE}\left[ \hat{u} + \hat{v} + \hat{p} \right] \\
        \mathcal{L}^{\text{fluid}}_{r_c}(\theta_1) &= \text{MSE}\left[ \frac{\partial \hat{u}}{\partial x} + \frac{\partial \hat{v}}{\partial y} \right] \\
        \mathcal{L}^{\text{fluid}}_{r_u}(\theta_1 ) &= \text{MSE}\left[ \frac{\partial \hat{u}}{\partial t} + \hat{u} \frac{\partial \hat{u}}{\partial x} + \hat{v} \frac{\partial \hat{u}}{\partial y} + \frac{1.0}{ \rho_{f}} \frac{\partial \hat{p}}{\partial x} - \mu  \left(\frac{\partial^2 \hat{u}}{\partial x^2} + \frac{\partial^2 \hat{u}}{\partial y^2}\right)  \right] \\
        \mathcal{L}^{\text{fluid}}_{r_v}(\theta_1 ) &= \text{MSE}\left[ \frac{\partial \hat{v}}{\partial t} + \hat{u} \frac{\partial \hat{v}}{\partial x} + \hat{v} \frac{\partial \hat{v}}{\partial y} + \frac{1.0}{ \rho_{f}} \frac{\partial \hat{p}}{\partial y} - \mu \left( \frac{\partial^2 \hat{v}}{\partial x^2} + \frac{\partial^2 \hat{v}}{\partial y^2} \right)  \right] \\
        \mathcal{L}_{\xi}(\theta_1 \cup \theta_2) &=  \text{MSE}\left[ \hat{u}(\xi(s,t)) - \Tilde{u}(\xi(s,t)) \right] + \text{MSE}\left[ \hat{v}(\xi(s,t)) - \Tilde{v}(\xi(s,t)) \right], \label{eq:m3_v_coupling}\\
        \mathcal{L}_{s}( \theta_2) &=   \text{MSE}\left[ n_x *\frac{\partial \Tilde{p}(\xi(s,t))}{\partial x} +  n_y *\frac{\partial \Tilde{p}(\xi(s,t))}{\partial y} \right] \notag
    \end{align} 
\end{subequations}
\noindent where $\mathcal{L}_{phy}^{fluid}$, $\mathcal{L}_{s}$  are the physics losses of the fluid and the structure, respectively. $\mathcal{L}_{left} $,  $\mathcal{L}_{right}$, $\mathcal{L}_{up}$,  $\mathcal{L}_{bottom,right}$,  $\mathcal{L}_{{u}_0}$, are the left, right, up, bottom and initial losses respectively.
In Eq.~\ref{eq:m3_v_coupling}, $\hat{u}, \hat{v}$ represent the predicted velocity components from the Eulerian (fluid) network at the interface, while $\Tilde{u}, \Tilde{v}$ are the velocity components predicted by the Lagrangian (interface) network at the same points. This term penalizes any differences between the predicted fluid and structure velocities, enforcing a no-slip boundary condition at the interface.

\subsection{Eulerian-Lagrangian with Learnable Activation (EL-L)}

This architecture combines the benefits of domain specialization (Section~\ref{sec:EL}) with adaptive activation functions (Section~\ref{sec:B-L}). As illustrated in Fig.~\ref{fig:EL}, both Eulerian and Lagrangian subnetworks employ learnable B-spline+SiLU activations instead of fixed Tanh functions, enabling localized adaptation to sharp gradients near fluid-structure boundaries while maintaining domain-specific optimization.
The mathematical formulation follows Section~\ref{sec:EL} for domain decomposition and Section~\ref{sec:B-L} for activation functions, providing the synergistic benefits of specialized network capacity allocation and locality-aware representation for capturing multi-scale FSI phenomena.


\section{Results and Discussion}

\subsection{Experimental Setup}

\label{sec:models}
\begin{table*}[t]
\small
\centering
\caption{
Presents four models developed: the baseline (B) and proposed architectures (B-L, EL, and EL-L), with their configurations evaluated in this work.
}
\label{tab:settings}

\begin{tabular}{@{}
        l
        l
        r
        l
        c
        @{}
        }
        
    \hline
    \textbf{Model}& \textbf{} & \textbf{Structure}& \textbf{Network}  &\textbf{Parameters}\\
    
    \hline\noalign{\vskip 1ex}
 
    Baseline (Fig.~\ref{fig:B}) & B & \multirow{2}{*} {} & [3, 600, 600, 600, 3]   & 725404\\
    
    \hline \noalign{\vskip 1ex}
    
    Baseline with Learnable Activation (Fig.~\ref{fig:B-L}) & B-L &  &  [3, 100, 100, 100, 3]  & 206001\\

    \hline \noalign{\vskip 1ex}

    \multirow{2}{*}{Eulerian-Lagrangian (Fig.~\ref{fig:EL})} & \multirow{2}{*}{EL }&  Eulerian & [3, 350, 350, 350, 3] &  \multirow{2}{*}{253608}\\

    \cline{3-4} \noalign{\vskip 1ex}

    & &  Lagrangian & [3, 50, 50, 50, 3] & \\

    \hline \noalign{\vskip 1ex}

   \multirow{2}{*}{\makecell[l]{Eulerian-Lagrangian with Learnable \\ Activation (Fig.~\ref{fig:EL-L})}} & \multirow{2}{*}{EL-L}& Eulerian &  [3, 100, 100, 100, 3]  & \multirow{2}{*}{259002} \\

    \cline{3-4} \noalign{\vskip 1ex}

    & & Lagrangian & [3, 50, 50, 50, 3] &  \\
    
    \hline
\end{tabular}
\label{table:models}
\end{table*}

Four neural network architectures were implemented and evaluated: baseline (B), baseline with learnable activation (B-L), Eulerian-Lagrangian (EL), and Eulerian-Lagrangian with learnable activation (EL-L), as summarized in Table~\ref{table:models}. All models employed fully connected feedforward networks with Xavier normal initialization to ensure stable gradient flow during training. Input normalization to the range $[-1, 1]$ enhanced training stability for Tanh activations.

Training was standardized across all models to ensure fair comparison: 60,000 iterations, using PyTorch's Adam optimizer ($\beta_1 = 0.9$, $\beta_2 = 0.999$, and $\epsilon = 10^{-8}$), with a learning rate decay (step = 1,000 and rate = 0.99). 
A Sobol sequence generated initial training datasets, with mini-batch gradient descent (batch size = 128) applied to randomly selected subsets.
The training dataset comprised 0.005\% of fluid domain and 0.05\% of solid interface data, ensuring efficient yet representative sampling of boundary conditions and the solid interface. For additional information and implementation details, please refer to our GitHub source code~\cite{afrah2024pinn_ibm4fsi}.

Performance evaluation used the relative $L_2$-norm: $\text{RMSE} = \sqrt{\frac{1}{n} \sum_{i=1}^{n} (f_i - \hat{f}_i)^2} * 100\%$, where $\hat{y}$ and $y$ are the predicted and the reference solutions, respectively. Reference solutions are obtained from high-precision Immersed Boundary Adaptive Mesh Refinement (IBAMR) software simulations. 
All experiments are performed on an NVIDIA A100 machine with a single GPU with 40 GB of VRAM.


\begin{table*}[t]
\small
\centering
\caption{
The RMSE error (in \%) of the models B, B-L, EL, and EL-L, detailed in Table~\ref{tab:settings}, along with the percentage improvement.
$\Delta_{\text{B}\to\text{B-L}}$ and $\Delta_{\text{EL}\to\text{EL-L}}$: \% of improvement for models using learnable activation over fixed activation. 
$\Delta_{\text{B}\to\text{EL}}$ and $\Delta_{\text{B}\to\text{EL-L}}$: \% of improvement for the Eulerian-Lagrangian over the baseline.
$\Delta_{\text{B}\to\text{EL-L}}$ shows that the EL-L model significantly outperforms the baseline, B.
}
\label{tab:RMSE_results}

\begin{tabular}{@{}
            l
            l!{\color{lightgray}\vline}
            c 
            c 
            c
            c!{\color{lightgray}\vline}
            c
            c
            c
            c@{}}
            
    \hline \noalign{\vskip 1ex}
    
    & & \makecell[c]{B}$\downarrow$& 
    \makecell[c]{B-L}$\downarrow$& 
    \makecell[c]{EL}$\downarrow$& 
    \makecell[c]{EL-L}$\downarrow$& 
    $\Delta_{\text{B}\to\text{B-L}}\uparrow$& 
    $\Delta_{\text{EL}\to\text{EL-L}}\uparrow$& 
    $\Delta_{\text{B}\to\text{EL}}\uparrow$& 
    $\Delta_{\text{B}\to\text{EL-L}}\uparrow$\\
    
    \hline \noalign{\vskip 1ex}
    \multirow{3}{*}{\makecell{Fluid}} & $u_x$ & 3.65 & 2.45 & 2.73 & \textbf{2.42} & 32.9 & 11.4 & 25.2 & \textbf{33.7} \\
    & $v_y$ & 3.33 & 2.41 & 3.45 & \textbf{2.49} & 27.6 & 27.8 & -3.6 & \textbf{25.2} \\
    & $p$ & 5.43 & 5.88 & 6.96 & \textbf{4.12} & -8.3 & 40.8 & -28.2 & \textbf{24.1} \\
    \hline \noalign{\vskip 1ex}
    \multirow{3}{*}{\makecell{Structure}} & $u_x$ & 1.99 & 1.14 & 0.74 & \textbf{0.24} & 42.7 & 67.6 & 62.8 & \textbf{87.9} \\
    & $v_y$ & 2.69 & 1.78 & 0.65 & \textbf{0.23} & 33.8 & 64.6 & 75.8 & \textbf{91.4} \\
    & $p$ & 12.90 & 10.00 & 5.19 & \textbf{2.39} & 22.5 & 53.9 & 59.8 & \textbf{81.5} \\
    \hline \noalign{\vskip 1ex}
\end{tabular}
\end{table*}

\begin{figure*}[htb]
    \centering
    \includegraphics[width=0.7\columnwidth, trim={0cm 0cm 0cm 0cm}, clip]{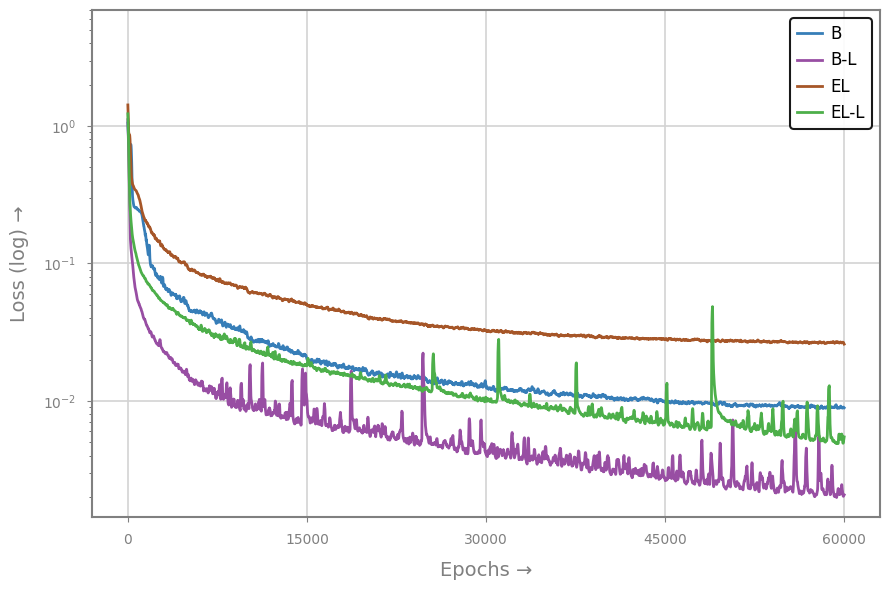} 
    \vspace{-0.2cm} 
    \caption{Training convergence for the baseline (B), the learnable baseline (B-L), the Eulerian-Lagrangian (EL), and the Learnable Eulerian-Lagrangian (EL-L) models over 60,000 epochs. These models correspond to the architectural and activation configurations detailed in Table~\ref{tab:settings}.
    }
   \label{fig:loss_history}
\end{figure*}

\begin{figure*}[ht]
\centering

\begin{tabular}{cc}  
    \includegraphics[width=0.45\columnwidth, trim={0cm 0cm 0cm 0cm}, clip]{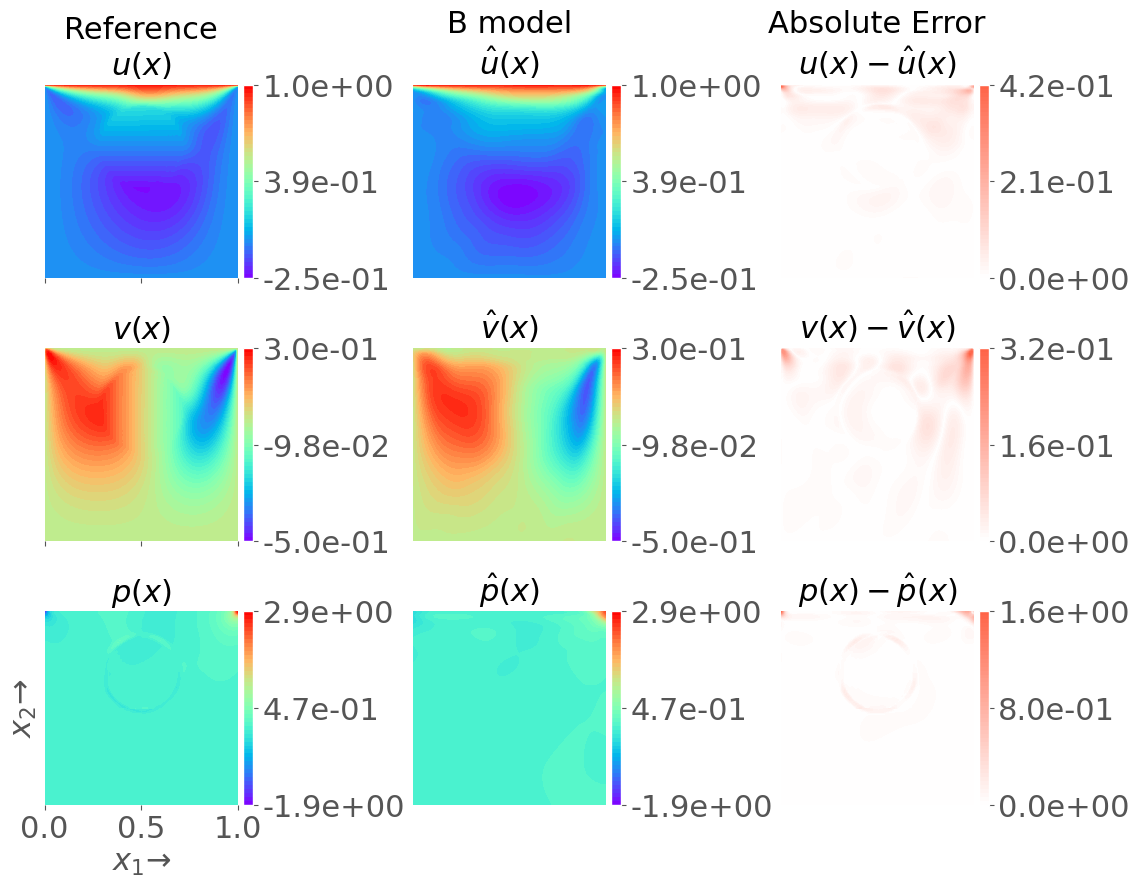} &
    \includegraphics[width=0.45\columnwidth, trim={0cm 0cm 0cm 0cm}, clip]{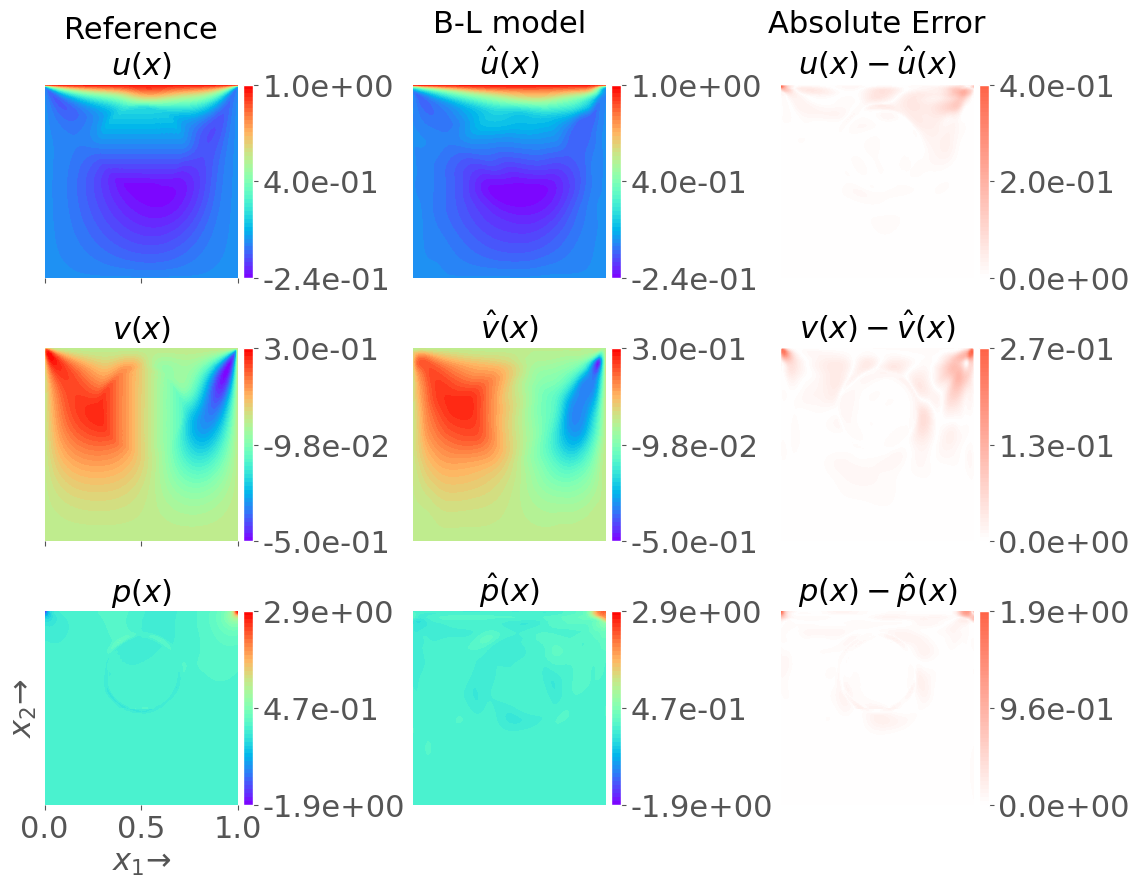} \\
 
    \scriptsize{(a) B (baseline)} & \scriptsize{(b) B-L (baseline enhanced with learnable activation)}\\
    \includegraphics[width=0.45\columnwidth, trim={0cm 0cm 0cm 0cm}, clip]{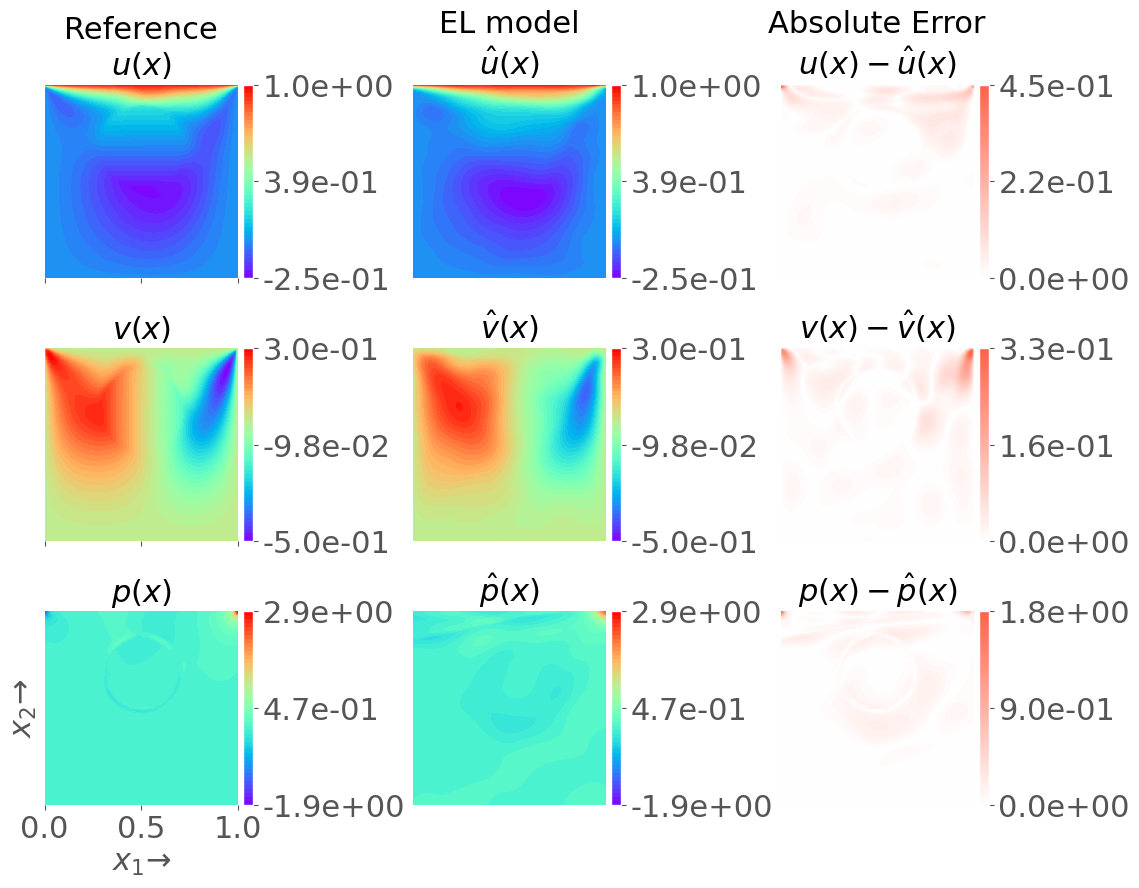} &
    \includegraphics[width=0.45\columnwidth, trim={0cm 0cm 0cm 0cm}, clip]{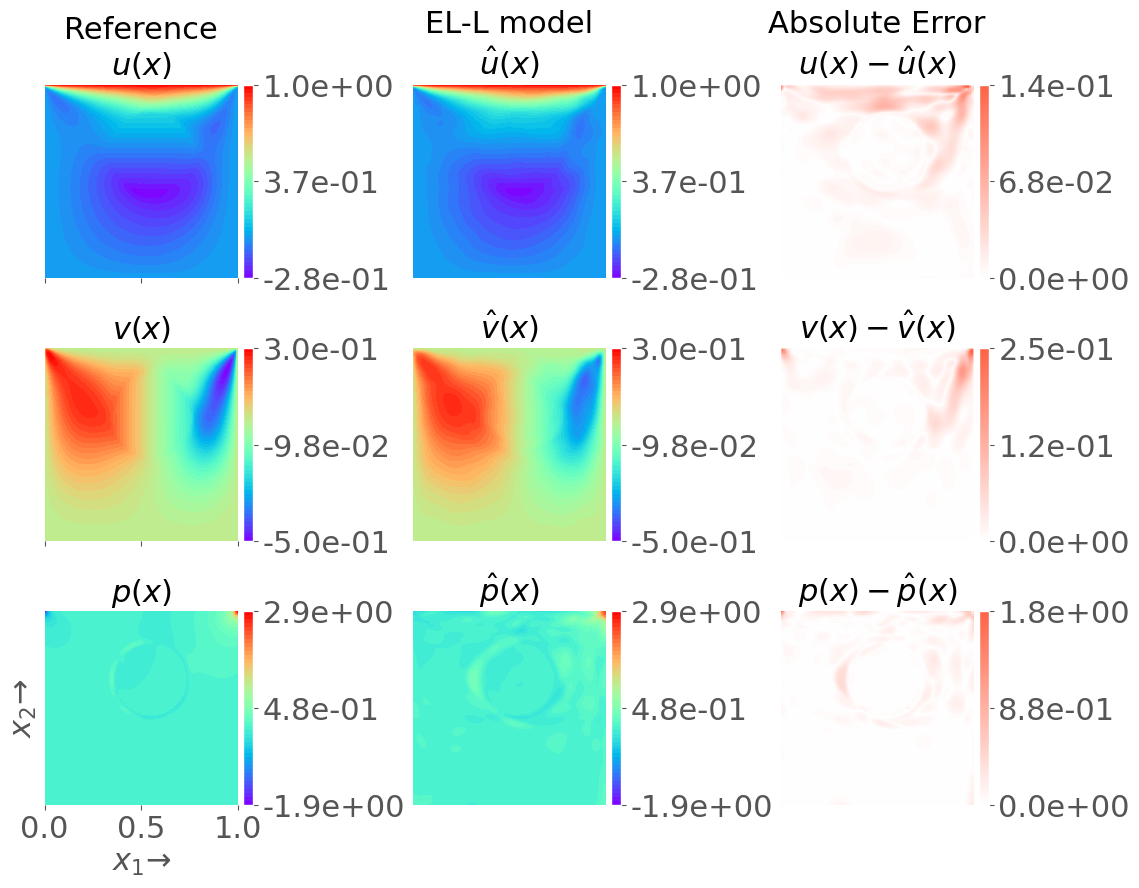} \\
   \scriptsize{(c) EL (Eulerian-Lagrangian)} & \scriptsize{(d) EL-L (Eulerian-Lagrangian with learnable activation)}
\end{tabular}

\caption{
    Contour plots (illustrating the Eulerian velocities $u_x, v_y$ and the pressure $p$ at steady state) for the solution of the selected FSI problem. The rows, from top to bottom, show the model-predicted solution, the reference CFD solution, and the point-wise absolute errors between the two solutions, respectively.
}
\label{fig:tricontourf_99}  
\end{figure*}

\begin{figure*}[!htb]
\centering
\begin{tabular}{@{}c m{0.70\columnwidth}@{}} 
    
  \rotatebox[origin=c]{90}{\tiny{B}}& 
  \includegraphics[width=1\linewidth, trim={0cm 0cm 0cm 0cm}, clip]{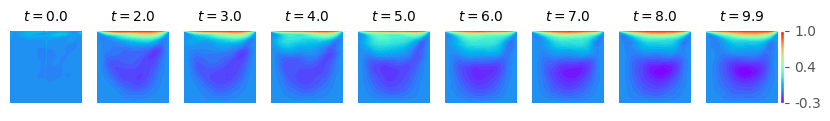} \\[-0.2cm] 
  
  \rotatebox[origin=c]{90}{\tiny{B-L}}& 
  \includegraphics[width=1\linewidth, trim={0cm 0cm 0cm 0.55cm}, clip]{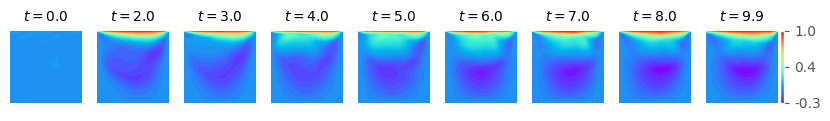} \\[-0.2cm]

  \rotatebox[origin=c]{90}{\tiny{EL}}& 
  \includegraphics[width=1\linewidth, trim={0cm 0cm 0cm 0.55cm}, clip]{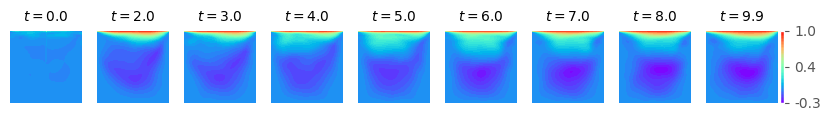} \\[-0.2cm]

  \rotatebox[origin=c]{90}{\tiny{EL-L}}& 
  \includegraphics[width=1\linewidth, trim={0cm 0cm 0cm 0.55cm}, clip]{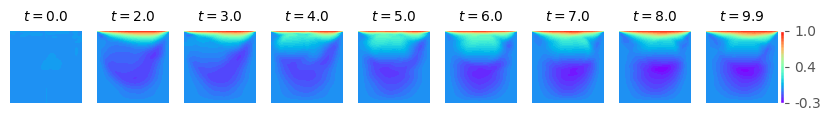} \\[-0.2cm]

  \rotatebox[origin=c]{90}{\tiny{Reference}}& 
  \includegraphics[width=1\linewidth, trim={0cm 0cm 0cm 0.55cm}, clip]{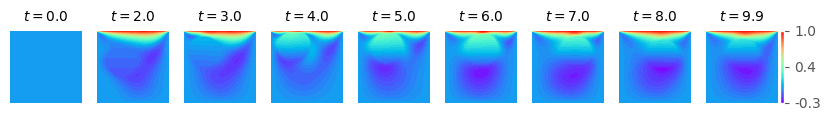} \\
    
\end{tabular}
\caption{
Comparison of the horizontal velocity component ($u_x$) predicted by models B, B-L, EL, and EL-L against the reference CFD solution at selected time steps on the Eulerian fluid domain.
}
\label{fig:eulerian_u}  
\end{figure*}

\begin{figure*}[htb]
\centering
\begin{tabular}{@{}c m{0.70\columnwidth}@{}}

  \rotatebox{90}{\tiny{B}}& 
  \includegraphics[width=1\linewidth, trim={0cm 0cm 0cm 0cm}, clip]{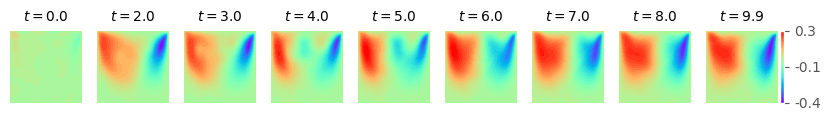} \\[-0.2cm]

  \rotatebox{90}{\tiny{B-L}}& 
  \includegraphics[width=1\linewidth, trim={0cm 0cm 0cm 0.55cm}, clip]{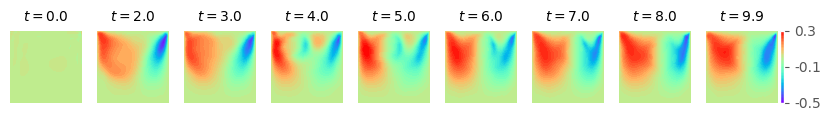} \\[-0.2cm]

  \rotatebox{90}{\tiny{EL}}& 
  \includegraphics[width=1\linewidth, trim={0cm 0cm 0cm 0.55cm}, clip]{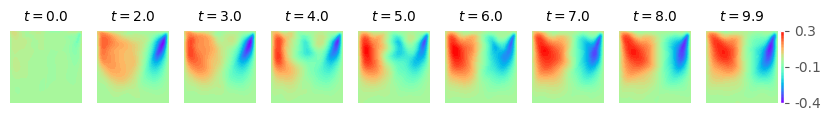} \\[-0.2cm]

  \rotatebox{90}{\makecell{\tiny{EL-L}}}& 
  \includegraphics[width=1\linewidth, trim={0cm 0cm 0cm 0.55cm}, clip]{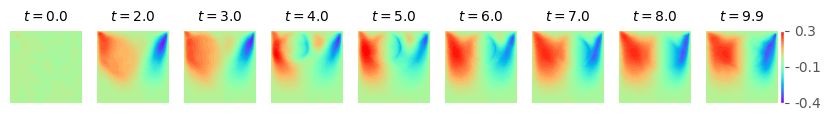} \\[-0.2cm]
  
  \rotatebox{90}{\tiny{Reference}}& 
  \includegraphics[width=1\linewidth, trim={0cm 0cm 0cm 0.55cm}, clip]{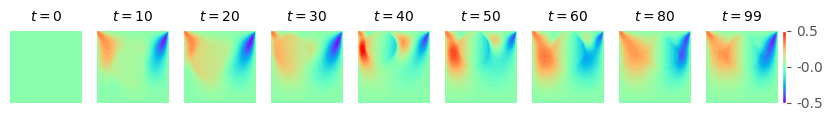} \\

\end{tabular}
\caption{Comparison of the vertical velocity component ($v_x$) predicted by models B, B-L, EL, and EL-L against the reference CFD solution at selected time steps on the Eulerian fluid domain.}
\label{fig:eulerian_v}  
\end{figure*}

\begin{figure*}[htb]
\centering

\begin{tabular}{@{}c m{0.70\columnwidth}@{}}

  \rotatebox{90}{\tiny{B}}& 
  \includegraphics[width=1\linewidth, trim={0cm 0cm 0cm 0cm}, clip]{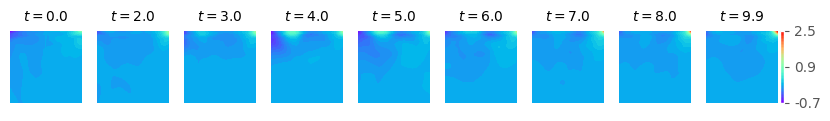} \\[-0.2cm]
  
  \rotatebox{90}{\tiny{B-L}}& 
  \includegraphics[width=1\linewidth, trim={0cm 0cm 0cm 0.55cm}, clip]{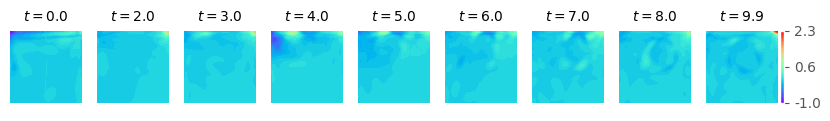} \\[-0.2cm]
  
  \rotatebox{90}{\tiny{EL}}& 
  \includegraphics[width=1\linewidth, trim={0cm 0cm 0cm 0.55cm}, clip]{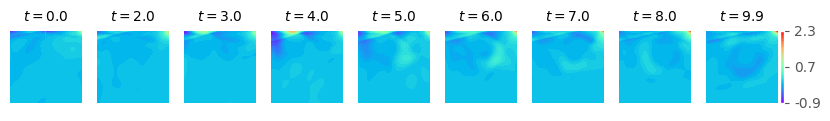} \\[-0.2cm]
  
  \rotatebox{90}{\tiny{EL-L}}& 
  \includegraphics[width=1\linewidth, trim={0cm 0cm 0cm 0.55cm}, clip]{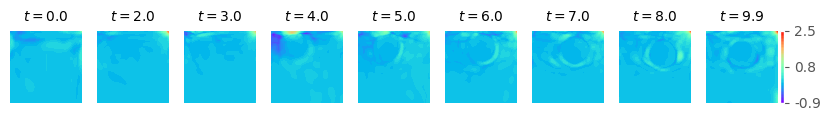} \\[-0.2cm]
  
  \rotatebox{90}{\tiny{Reference}}& 
  \includegraphics[width=1\linewidth, trim={0cm 0cm 0cm 0.55cm}, clip]{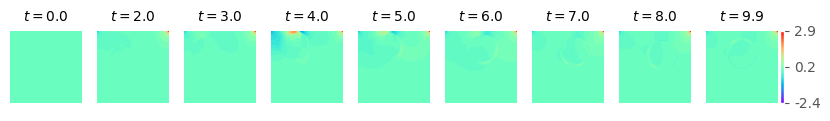} \\

\end{tabular}
\caption{Comparison of the pressure field ($p$) predicted by models B, B-L, EL, and EL-L against the reference CFD solution at selected time steps on the Eulerian fluid domain.}
\label{fig:eulerian_p}  
\end{figure*}




\begin{figure*}[!htb]
\centering
    \includegraphics[width=0.72\columnwidth, trim={0cm 0cm 0cm 0cm}, clip]{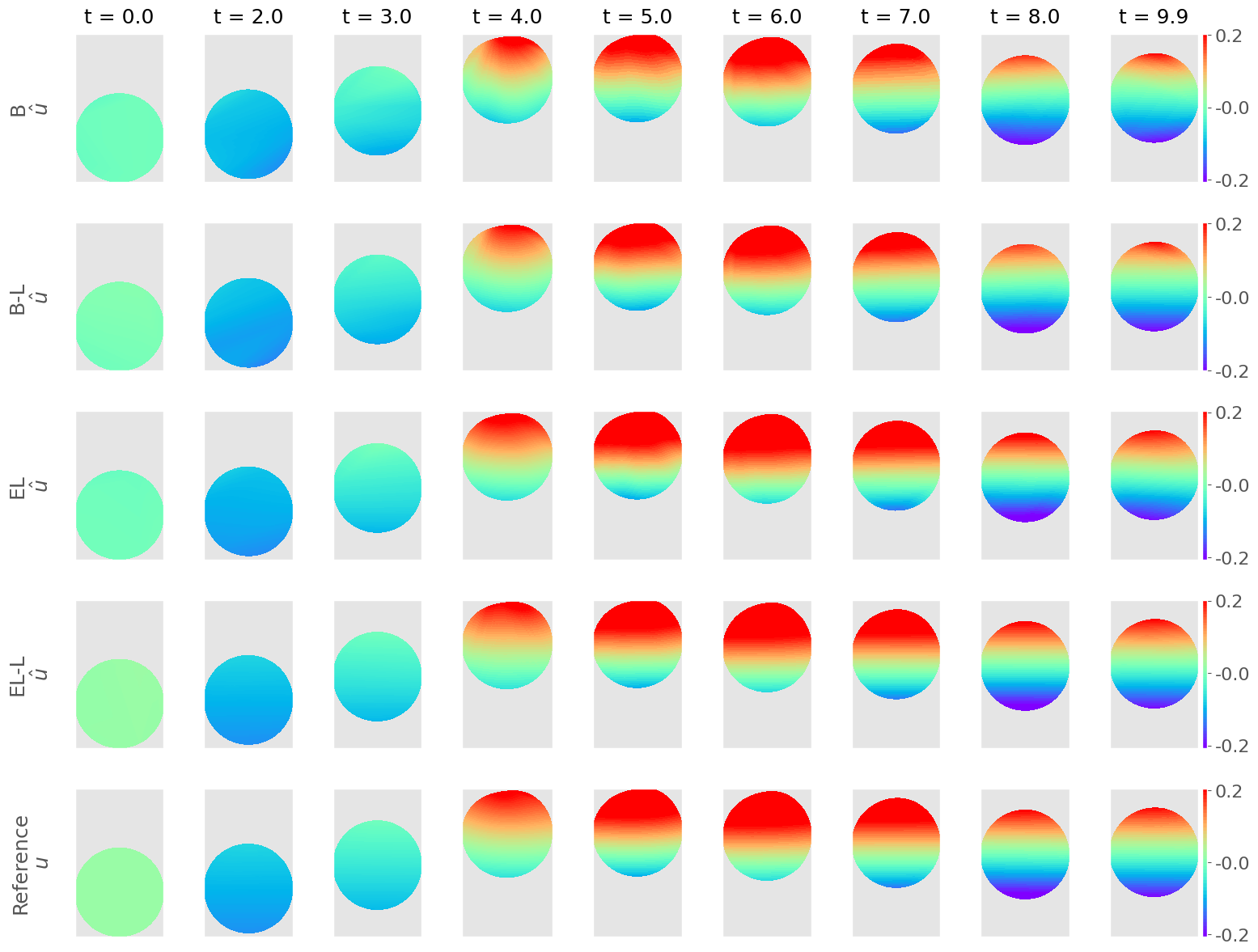} 
    \caption{
    Temporal changes in the horizontal velocity component ($u_x$) at the structure domain are presented for models B, B-L, EL, and EL-L compared to the reference CFD solution.
    }
\label{fig:structure_u}  
\end{figure*}

\begin{figure*}[!htb]
\centering
    \includegraphics[width=0.72\columnwidth, trim={0cm 0cm 0cm 0cm}, clip]{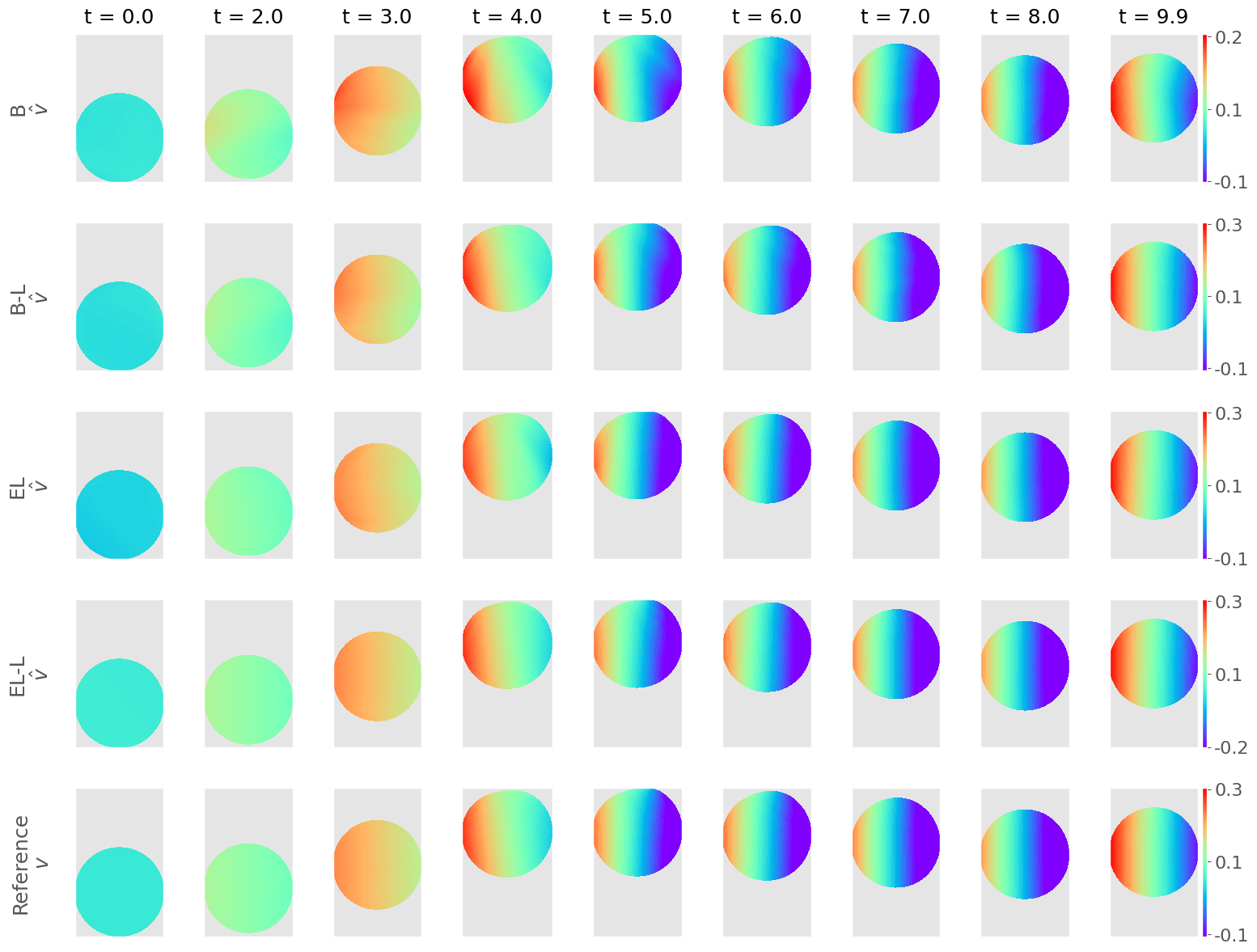} 
    \caption{
    Temporal changes in the vertical velocity component ($v_y$) at the structure domain are presented for models B, B-L, EL, and EL-L compared to the reference CFD solution.
    }
\label{fig:structure_v}  
\end{figure*}

\begin{figure*}[!htb]
\centering
    \includegraphics[width=0.72\columnwidth, trim={0cm 0cm 0cm 0cm}, clip]{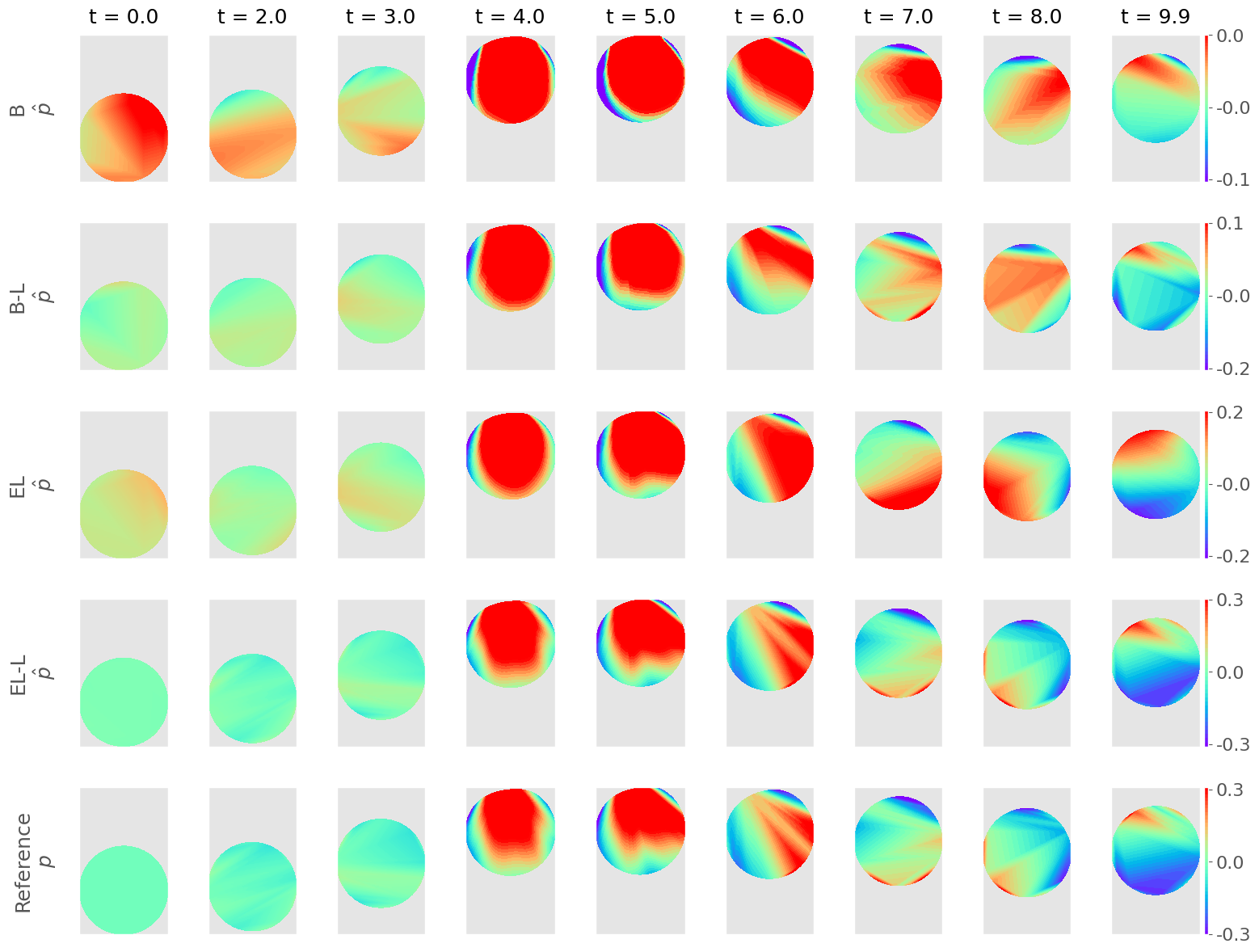} 
    \caption{
    Temporal changes in the pressure field ($p$) at the structure domain are presented for models B, B-L, EL, and EL-L compared to the reference CFD solution.
    }
\label{fig:structure_p}  
\end{figure*}


\subsection{Experimental Results}

Table~\ref{tab:RMSE_results} presents RMSE errors (in \%) for velocity and pressure predictions in both the fluid and structure domains. Three key trends emerge from these results:

\textbf{(a) Advantage of domain decomposition.}
Eulerian-Lagrangian architectures (EL and EL-L) achieve substantial improvements over unified baseline models, with performance gains varying by domain. In the fluid domain, EL shows mixed results compared to B, with 25\% improvement in $u_x$ but degraded performance in $v_y$ (-3.6\%) and pressure (-28\%), indicating that domain decomposition alone requires careful tuning. However, EL-L consistently outperforms B across all fluid metrics (25-34\%) velocity improvements, 24\% pressure improvement). In the structure domain, both EL and EL-L demonstrate better performance: EL reduces velocity errors by 62-76\% and pressure error by 60\% (12.9\% to 5.19\%) relative to B, while EL-L achieves even greater improvements of 88-91\% for velocity and 81\% for pressure prediction. This performance pattern reflects the specialized networks' ability to focus on domain-specific physics without parameter interference, with benefits most pronounced in the structurally complex interface region.

\textbf{(b) Impact of learnable activations.}
Models with learnable B-spline+SiLU activation functions (B-L and EL-L) consistently outperform their Tanh-based counterparts (B and EL). In the fluid domain, B-L reduces velocity errors by 27-33\% compared to B, while EL-L achieves velocity improvements of 11-28\% over EL. However, fluid domain pressure prediction shows contrasting behaviors: B-L exhibits slight deterioration (5.43\% to 5.88\%), whereas EL-L demonstrates better pressure improvement (6.96\% to 4.12\%, a 41\% reduction) over EL.
In the structure domain, learnable activations provide consistent benefits across all variables: B-L achieves velocity improvements of 34-43\% and pressure reduction of 22.5\% compared to B, while EL-L delivers even more dramatic improvements with velocity error reductions of 65-68\% and pressure improvement of 54\% over EL, highlighting the benefits of combining domain specialization with adaptive activations.

\textbf{(c) Combined benefits.}
The EL-L model achieves better performance across all metrics, reducing structural pressure errors from 12.9\% (baseline) to 2.39\%, representing an 81.5\% improvement. Structural velocity errors decrease to just 0.24\% ($u_x$) and 0.23\% ($v_y$), demonstrating the combined effect of domain specialization and adaptive activation functions.

\textbf{Training convergence analysis.}
Fig.~\ref{fig:loss_history} illustrates training convergence over 60,000 epochs. Models with B-spline+SiLU activations (B-L and EL-L) demonstrate accelerated convergence, particularly during the first 20,000 epochs, due to their adaptive representation capabilities. However, these models exhibit increased oscillatory behavior at later epochs, attributed to the learnable spline coefficients introducing additional gradient variability.
On the other hand,  Eulerian-Lagrangian models (EL and EL-L) converge more slowly than baseline architectures due to the complexity of balancing additional coupled loss terms (Eq.~\ref{eq:m3_coupling}). Despite slower convergence, these models achieve better final accuracy than the coupled models.

\textbf{Steady-state performance.}
Fig.~\ref{fig:tricontourf_99} compares steady-state predictions across all models. The baseline model (B) struggles with vortical structures and sharp gradients near the solid interface, exhibiting notable deviations in both velocity and pressure fields. The introduction of learnable activations (B-L) significantly improves velocity component predictions, particularly in bulk flow regions where transient vortical features dominate, but shows limited improvement in pressure prediction.
Eulerian-Lagrangian models (EL and EL-L) offer improved accuracy at the interface. The EL-L predictions nearly overlap with CFD reference solutions for both velocity and pressure.

\textbf{Temporal evolution in fluid domain.}
Fig.~\ref{fig:eulerian_u}-\ref{fig:eulerian_p} present temporal comparisons for velocity components and pressure in the Eulerian domain. The baseline model exhibits increasing errors at later time steps when sharp gradients dominate. B-L shows marked improvement in velocity predictions but maintains pressure prediction challenges. EL and EL-L models consistently track reference solutions throughout the simulation, with EL-L achieving near-perfect agreement.

\textbf{Structural domain dynamics.}
Fig.~\ref{fig:structure_u}-\ref{fig:structure_p} demonstrate temporal evolution within the structure domain. Discrepancies between model predictions and reference solutions become pronounced at later time steps due to sharp gradients and interface dynamics. The baseline model shows the largest errors, particularly in capturing abrupt boundary changes. The EL model achieves substantial improvements, reducing structural velocity RMSEs to 0.74\% ($u_x$) and 0.65\% ($v_y$), while EL-L maintains close agreement throughout the simulation with velocity errors below 0.24\%.

\subsection{Discussion}

\noindent\textbf{Benefits of domain specialization.}
Unified network architectures (B and B-L) struggle with heterogeneous target variable distributions across fluid and structure domains, leading to parameter sharing conflicts and compromised accuracy. The statistical analysis in Table~\ref{tab:ibm_dist} reveals that pressure variability at interfaces significantly exceeds bulk fluid regions, while velocity fields exhibit opposite trends. This fundamental mismatch necessitates domain-specific optimization strategies.
Our Eulerian-Lagrangian architectures address this challenge by decoupling fluid and structural predictions into specialized networks. This approach enables each subnetwork to optimize for domain-specific physics: the fluid network focuses on continuous flow fields governed by Navier-Stokes equations, while the interface network specializes in sharp gradients and boundary conditions. This specialization eliminates parameter interference and allows optimal capacity allocation for each domain's challenges.

\noindent\textbf{Role of activation function.}  
Learnable B-splines+SiLU activations outperform the fixed Tanh functions in capturing sharp interface dynamics and high-gradient regions. B-splines provide local adaptability through compact support and learnable control points, enabling precise adjustments in complex areas. SiLU activation enhances global expressivity with smooth, responsive nonlinearity that stabilizes training and captures overall flow patterns.
While Tanh functions offer advantages for smooth, low-frequency components, they struggle with sharp interfaces and localized features. The hybrid B-spline+SiLU approach effectively balances local precision with global coherence, proving essential for FSI problems with multi-scale solution features.

\noindent \textbf{Additional considerations.}
Beyond network architecture and activation functions, we explored various dynamic PINN loss-balancing schemes, e.g., Gradient Statistics approach~\cite{wang2021understanding}, Self-Adaptive (SA) method~\cite{mcclenny2023self}, and Residual-Based Attention (RBA)~\cite{anagnostopoulos2024residual}.
These methods proved ineffective for FSI applications. For instance, the gradient statistics method, in particular, encountered significant difficulties with  ``noisy'' loss terms associated with coupling equations. These terms occasionally trained well in all models, even under low-gradient conditions, likely because they contain specific frequencies that are learned faster than others~\cite{rahaman2019spectral, cao2019towards}.
The low gradients led the learnable methods to assign larger weights to these terms, creating an imbalance that hindered convergence rather than improving it.

\noindent \textbf{Limitations and future work.}
Current evaluation is limited to a single case study due to scarce open FSI datasets with moving boundary formulations. We are collaborating with CFD experts to develop comprehensive benchmark datasets for community use. Future work will incorporate force-coupling terms to improve momentum conservation and interfacial accuracy, as well as extend evaluation to higher-dimensional problems with complex geometries.


\section{Conclusion}

Our work establishes that architectural alignment with underlying physics is fundamental to successful PINN implementation for fluid-structure interaction problems. By decoupling Eulerian fluid and Lagrangian structural representations into specialized neural networks, we overcome the inherent limitations of unified architectures that force incompatible physics into shared parameter spaces. 

Our evaluation of 2D cavity flow with a moving disc shows that this domain-specific approach, combined with learnable B-spline and SiLU activations, reduces structural pressure errors from 12.9\% to 2.39\%. This significant improvement is crucial for fluid-structure interaction (FSI) applications, where the accuracy of the interface directly affects the overall fidelity of the solution.
The performance gap between unified and decoupled architectures reveals a broader principle for physics-informed learning: neural network design must reflect the mathematical structure of the governing equations. In FSI problems, the Eulerian-Lagrangian formulation naturally separates smooth bulk flow physics from sharp interfacial dynamics. Our architecture mirrors this separation, allowing each subnetwork to optimize for its domain's characteristic features without compromise. The 24-91\% error reduction across velocity and pressure fields validates this physics-driven design approach.

Equally significant is the role of activation functions in capturing multi-scale phenomena. While fixed Tanh activations provide a stable global approximation, they fail to resolve the high-gradient features characteristic of fluid-structure interfaces. Learnable B-splines with SiLU activation achieve locality-aware representation, adapting their basis functions to concentrate resolution where needed (near moving boundaries) while maintaining efficiency in smooth regions. This adaptive capability proves essential for FSI problems where solution features span multiple length scales.

Overall, our research demonstrates the significant advantages of deploying domain-specialized architectures and learnable activation functions in effectively addressing FSI challenges using physics-informed machine learning.

\keywords{
Fluid Structure Interaction (FSI) \and Immersed Boundary Method (IBM) \and Moving Boundary \and Physics Informed Neural Networks (PINN)\and Activation
}

\bibliographystyle{unsrt}  
\bibliography{references}  


\end{document}